\documentclass[letterpaper]{article} 
\usepackage{aaai25}  
\usepackage{times}  
\usepackage{helvet}  
\usepackage{courier}  
\usepackage[hyphens]{url}  
\usepackage{graphicx} 
\usepackage{natbib}  
\usepackage{caption} 
\usepackage{algorithm}
\usepackage{algorithmic}

\usepackage{newfloat}
\usepackage{listings}
\DeclareCaptionStyle{ruled}{labelfont=normalfont,labelsep=colon,strut=off} 
\floatstyle{ruled}
\newfloat{listing}{tb}{lst}{}
\floatname{listing}{Listing}
\title{GIM: A Million-scale Benchmark for Generative Image Manipulation Detection and Localization}
\author{
Yirui Chen\textsuperscript{\rm 1,3,\equalcontrib}  ,
Xudong Huang\textsuperscript{3,\equalcontrib
} , Quan Zhang \textsuperscript{2,3,\equalcontrib
} ,
Wei Li\textsuperscript{3} ,
Mingjian Zhu\textsuperscript{3},
Qiangyu Yan\textsuperscript{3},
Simiao Li \textsuperscript{3},
Hanting Chen\textsuperscript{3},
Hailin Hu\textsuperscript{3},
Jie Yang\textsuperscript{1},
Wei Liu\textsuperscript{1,\thanks{Corresponding authors.}},
Jie Hu\textsuperscript{3,\footnotemark[2]}
}
\affiliations{
    \textsuperscript{\rm 1} Shanghai Jiao Tong University\\
    \textsuperscript{\rm 2} Tsinghua University\\
    \textsuperscript{\rm 3} Huawei Noah’s Ark Lab\\
    \{chenyirui,weiliucv\}@sjtu.edu.cn    zhangqua22@mails.tsinghua.edu.cn \{huangxudong9,wei.lee,hujie23\}@huawei.com
}

\usepackage{bibentry}

\usepackage{algorithm}
\usepackage{algorithmic}

\usepackage{multirow}
\usepackage{bm}
\usepackage{makecell}
\usepackage{amsmath}
\usepackage{booktabs}
\usepackage{pifont}

\begin{document}

\maketitle

\begin{abstract}
The extraordinary ability of generative models emerges as a new trend in image editing and generating realistic images, posing a serious threat to the trustworthiness of multimedia data and driving the research of image manipulation detection and location (IMDL). However, the lack of a large-scale data foundation makes the IMDL task unattainable. In this paper, we build a local manipulation data generation pipeline that integrates the powerful capabilities of SAM, LLM, and generative models. Upon this basis, we propose the GIM  dataset, which has the following advantages: 1) Large scale,  GIM includes over one million pairs of AI-manipulated images and real images. 2) Rich image content, GIM encompasses a broad range of image classes. 3) Diverse generative manipulation, the images are manipulated images with state-of-the-art generators and various manipulation tasks. The aforementioned advantages allow for a more comprehensive evaluation of IMDL methods, extending their applicability to diverse images. We introduce the GIM benchmark with two settings to evaluate existing IMDL methods. In addition, we propose a novel IMDL framework, termed GIMFormer, which consists of a ShadowTracer, Frequency-Spatial block (FSB), and a Multi-Window Anomalous Modeling (MWAM) module. Extensive experiments on the GIM demonstrate that GIMFormer surpasses the previous state-of-the-art approach on two different benchmarks.
\end{abstract}

%

\section{Introduction}\label{sec:intro}

\begin{figure*}[t]
	\centering
	\includegraphics[width=0.75\textwidth]{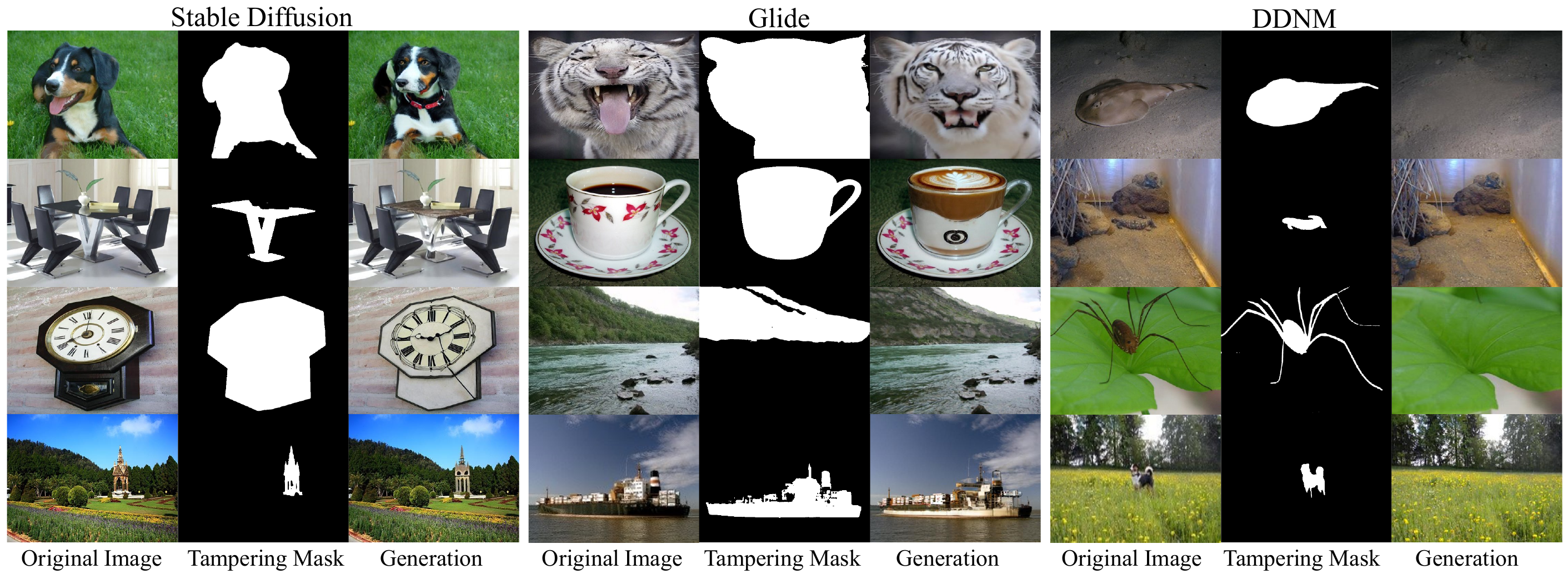}
	\caption{Example images from the GIM dataset. Our dataset includes images manipulated by three state-of-the-art generators: Stable-Diffusion, GLIDE, and DDNM. Three columns display authentic images, manipulation masks and manipulated images. }
	\label{fig:benchmark_vis}

\end{figure*}	
Images are one of the most essential media for information transmission in modern society and they are widely spread on public platforms such as news and social media. With the rapid advancement of generative methods~\cite{dhariwal2021diffusion, rombach2022high}, such natural information can be more easily manipulated for specific purposes such as tampering with an object or person. The image of this class is particularly convincing since its visual comprehensibility, leading to serious information security risks in social areas. 
Therefore, it is of utmost urgency to develop methods to detect whether an image is modified by generative models and identify the exact location of the manipulation. However, traditional image manipulation detection and location (IMDL) datasets~\cite{ dong2013casia, wen2016coverage} overlook the powerful generative models and generative IMDL~\cite{jia2023autosplice,guillaro2023trufor} datasets are limited with scale, it is thus not sufficient to comprehensively evaluate the performance of IMDL methods and benefit the IMDL community.

To this end, we propose a million-level generative-based IMDL dataset, termed GIM dataset, to provide a reliable database for AI Generated Content (AIGC) security.
GIM leverages the generative models~\cite{ho2020denoising,yin2025multi} and SAM~\cite{Kirillov_2023_ICCV}, with the images in  ImageNet~\cite{deng2009imagenet} and VOC~\cite{everingham2010pascal}  as the input. SAM is utilized to locate the tampering region, and generative models paint the reasonable content in the tampering area.
GIM contains a total of over one million generative manipulated images. To develop an appropriate benchmark scale, we explore the impact of different amounts of training manipulated data. The final benchmark contains about 320k manipulated images with their tampering masks for training and testing. To simulate the real situation, we investigate the effect of degradation and apply random degradation to these benchmark images. Based on this GIM benchmark, the IMDL methods are evaluated and benchmarked. Overall, GIM possesses the following advantages: 1) GIM has a large and reasonable data scale, including rich image categories and contents. 2) GIM contains various generative manipulation models and tasks. 3) GIM proposes two settings for verifying the performance and generalization ability of IMDL methods.

Existing methods emphasize traditional image manipulations also known as ``cheapfakes''. However, generative manipulations introduce lethal alterations in content with no apparent frequency or structural inconsistency. 
To address the above issue, we introduce GIMFormer, a transformer-based framework for generative IMDL.
The ShadowTracer is designed to embed the nuanced artifacts inherent in generative tampering and serves as the prior information. The Frequency-Spatial Block captures the manipulation clues in the frequency and spatial domains. Furthermore, the Multi Windowed Anomalous Modelling module captures local inconsistencies at different scales to refine the features. GIMFormer extracts features from both the RGB and learned tampering trace maps, capturing details from both the frequency and spatial domains while modeling inconsistencies at different scales for precise manipulation detection and localization. We conduct experiments on our proposed GIM. Both the qualitative and quantitative results demonstrate that GIMFormer can outperform previous state-of-the-art methods. In summary, our main contributions are as follows:

\begin{itemize}
	\item 
    We build a data generation pipeline and construct a large-scale dataset for IMDL tasks. 
    \item We investigate the impact of data scales and degradation, constructing a comprehensive benchmark with two evaluation settings for IMDL methods evaluation.
	\item We propose a framework named GIMFormer for generative IMDL task. Extensive experiments demonstrate that GIMFormer achieves state-of-the-art performance.
\end{itemize}

\section{Related Work}
\subsection{Image Forensic Datasets}
Early datasets~\cite{ng2004data} primarily focus on one type of manipulation.  
CASIA ~\cite{dong2013casia} first incorporates multiple types of manipulations with forged images manually crafted using Adobe Photoshop. The Wild Web dataset~\cite{zampoglou2015detecting} collects forged images from the Internet, surpassing previous datasets in the scale.  NIST \cite{guan2019mfc} dataset provides an extensive collection of datasets serving as a crucial standard for assessing media tampering detection methods.
IMD2020~\cite{novozamsky2020imd2020} provides locally manipulated images generated through manual operations or random slicing and online images without obvious manipulation traces.
Recently, HiFi-IFDL~\cite{guo2023hierarchical} constructs a hierarchical fine-grained dataset containing some representative forgery methods.
AutoSplice~\cite{jia2023autosplice} leverages large-scale language-image models like DALL-E2 ~\cite{ramesh2021zero} to facilitate automatic image editing and generation. CocoGlide~\cite{guillaro2023trufor} contains images manipulated by the diffusion model. Besides, other datasets focus on facial manipulations~\cite{rossler2019faceforensics++,guarnera2022face} or entirely synthesized images\cite{zhu2023genimage,yan2024sanity,verdoliva2022}.

The datasets mentioned above have limitations such as small data sizes and limited manipulation techniques. Recent advances in generative models have demonstrated remarkable manipulation abilities. 
Leveraging these models, we introduce GIM, a large-scale dataset that incorporates various recent generative manipulation techniques.

\begin{figure*}[!t]
	\centering
	\includegraphics[width=0.88\textwidth]{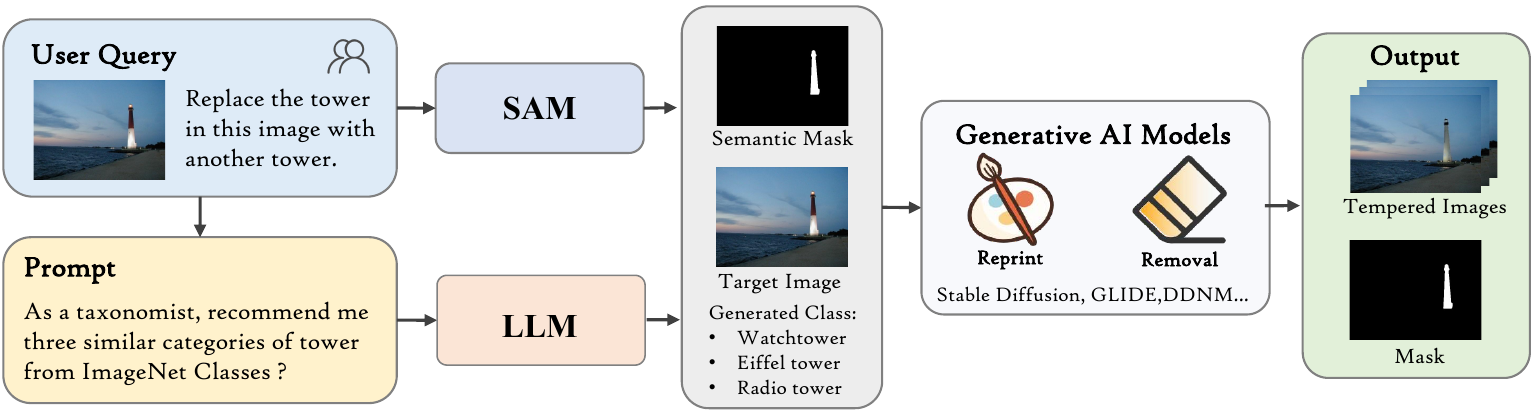}
	\caption{An overview of the dataset generation. Given the original image and a user query (classification attribution or mouse input), the manipulation mask is extracted using SAM. Tampering prompts are then organized with LLM by combining replacement classes. The final generations are produced by generative models with the image, tampering mask and prompts.}
	\label{fig:benchmark_pipeline}

\end{figure*}

\subsection{Image Manipulation Detection and Localization}
Early studies on natural image manipulation localization mainly focus on detecting a specific type of manipulation~\cite{cozzolino2015splicebuster,yin2023information}. 
Due to the exact manipulation type is unknown in real-world scenarios, most state-of-the-art methods~\cite{liu2022pscc,ying2023learning} primarily concentrate on general manipulation. 
MVSS-Net~\cite{dong2022mvss} uses a two-stream CNN to extract noise features and employs dual attention to fuse their outputs. PSCC-Net~\cite{liu2022pscc} extracts hierarchical features in a top-down manner and it detects manipulations in a bottom-up manner.   Motivated by the power of transformer\cite{zhang2024distilling,zhang2024can,lu2024deep,lu2024cricavpr,chen2023fastc}, ObjectFormer~\cite{wang2022objectformer} uses object prototypes to model object-level consistencies and find patch-level inconsistencies to detect the manipulation.
However, these methods focus on ``cheapfake'' detection and encounter challenges when applied to generative manipulation. Certain works~\cite{huang2022fakelocator,chai2020makes,liu2024cdngpfastscalablecontinual} localize manipulations using generative models but mainly focus on human faces.

In this work, we focus on the generative IMDL task.  We leverage a deep network to embed the subtle artifacts inherent in generative tampering as a prior trace map. A dual network fuses the trace map and RGB image, combining frequency and spatial information and capturing pixel-level inconsistencies to detect and locate the manipulation.

\section{Dataset and Benchmark Construction}
In this section, We propose an automatic pipeline to generate manipulated images from unannotated data. Leveraging this pipeline, we construct the comprehensive large-scale GIM dataset. To build a reasonable benchmark, we conduct preliminary experiments focusing on two key aspects: data scale and image degradation. First, we investigate the impact of the training data scale to determine an appropriate size for the GIM benchmark.  Second, to reveal real-world scenarios, the manipulated and original images are subjected to three random degradations. Finally, the GIM benchmark comprises over 320k manipulated images paired with authentic images for algorithm training and evaluation. The example images are shown in Figure \ref{fig:benchmark_vis} with their original images and tampering masks. Finally, we introduce the criteria and settings used to evaluate the IMDL methods.

\begin{table*}[!t]
\centering
\begingroup
\fontsize{7pt}{8pt}\selectfont
\setlength{\tabcolsep}{2pt}
\renewcommand{\arraystretch}{1.0} 
\begin{tabular}{cccccccc}
\toprule[1.5pt]
\multirow{2}{*}{Dataset} & \multirow{2}{*}{Image Content} & \multirow{2}{*}{Image Size} & \multicolumn{2}{c}{Num. Images} & \multicolumn{3}{c}{Manipulations Category} \\
                         &                                &                             & \# Authentic Image & \# Forged Image & Traditional & Gen.Reprint & Gen.Removal \\ \midrule
FaceForensics++~\cite{rossler2019faceforensics++} & Face  & 480p-1080p        & 1,000  & 4,000  & \ding{55} & \ding{51} & \ding{55} \\
DFDC~\cite{dang2020detection}                      & Face  & 240p-2160p        & 19,154 & 100,000 & \ding{55} & \ding{51} & \ding{55} \\
DeeperForensics~\cite{jiang2020deeperforensics}    & Face  & 1080p             & 50,000 & 10,000  & \ding{55} & \ding{51} & \ding{55} \\
Columbia Gray~\cite{ng2004data}                   & General & 128 $\times$ 128   & 933    & 912     & \ding{51} & \ding{55} & \ding{55} \\
CASIA V2.0 ~\cite{dong2013casia}                  & General & 320 $\times$ 240-800 $\times$ 600 & 7,200  & 5,123   & \ding{51} & \ding{55} & \ding{55} \\
IMD2020~\cite{novozamsky2020imd2020}              & General & 1062 $\times$ 866  & 35,000 & 35,000  & \ding{51} & \ding{55} & \ding{55} \\
Coverage~\cite{wen2016coverage}                   & General & 400 $\times$ 486   & 100    & 100     & \ding{51} & \ding{55} & \ding{55} \\
AutoSplice  ~\cite{jia2023autosplice}             & General & 256 $\times$ 256 - 4232 $\times$ 4232 & 3,621  & 2,273   & \ding{55} & \ding{51} & \ding{55} \\
HiFi-IFDL~\cite{guo2023hierarchical}              & General & -                  & -      & 1,000,000 (200,000) & \ding{51} & \ding{51} & \ding{55} \\
CocoGlide~\cite{guillaro2023trufor}               & General & 256$\times$256     & -      & 512     & \ding{55} & \ding{51} & \ding{55} \\
GIM                                              & General & 64$\times$64-6000$\times$3904 & 1,140,000 & 1,140,000 & \ding{55} & \textbf{\ding{51}} & \textbf{\ding{51}} \\ 
\bottomrule[1.5pt]
\end{tabular}
\caption{Summary of image manipulation datasets. Image content, image sizes, and manipulation techniques are reported. HIFI-IFDL includes mainly entirely synthesized images with a small portion of traditional manipulation or face image manipulation.}
\endgroup
\label{GIM:compare}
\end{table*}

\subsection{Data Generation Pipeline}\label{sec:data}
Benefiting from open-source projects~\cite{kirillov2023segment,ren2024grounded},  we develop our data generation pipeline. 
Figure \ref{fig:benchmark_pipeline} illustrates the overall process.  Our pipeline includes two types of manipulation: reprinting the target class using a generative inpainting method Stable Diffusion~\cite{rombach2022high} and GLIDE ~\cite{nichol2021glide} or removing the destination region using a specific removal method DDNM~\cite{wang2022zero}. Given an arbitrary image, the process begins by extracting the local manipulation mask using a zero-shot segmentation network SAM, guided by either classification attributes or user queries. For reprint tampering, the image category is embedded in the replacement prompt and interacts with LLM~\cite{achiam2023gpt}, which returns an approximate category. The approximate category is then embedded into the inpainting prompt.  By combing with the original image, manipulation mask and inpainting prompt, generative models generate the reprint tampered result. For removal tampering, only the original image and manipulation mask are required for the model. 

The GIM dataset utilizes images from the ImageNet dataset ~\cite{deng2009imagenet} and VOC dataset ~\cite{everingham2010pascal}. These two datasets contain a large number of diverse images with wide category coverage, providing a reliable database and laying the foundation for future research.

\subsection{Analysis of the Benchmark Scale}\label{sec:scale}
With the data generation pipeline, we are qualified to generate large amounts of manipulated images. Nonetheless, blindly increasing the amount of data does not improve the algorithm performance, while it may lead to data redundancy. Therefore, taking data generated by the Stable Diffusion as an example, we explore the influence of data volume and category volume on baseline classification~\cite{he2016deep} and segmentation algorithms~\cite{xie2021segformer}. Training sets vary in the scale, while the validation uses the same test set. As shown in Table \ref{tab:benchmark_scale}, the metrics of classification and segmentation are improved as the dataset scale increases. When the scale reaches 180K, the algorithm performance almost saturates. All the metrics remain almost unchanged with either the image class number or the data scale increased. Experiments demonstrate that increasing the amount of data or category brings negligible benefits when the data tends to be saturated. According to the analysis, the GIM benchmark uses 100 labels from ImageNet to generate tampered images for training and employs all the test sets from ImageNet and VOC for evaluation.

\begin{figure*}[!t]
    \begin{minipage}[t]{0.5\textwidth}
        \centering

        \vspace{-12mm}
       \renewcommand{\arraystretch}{1.04} 

        {\fontsize{7pt}{8pt}\selectfont 
        \begin{tabular}{c|c|c|ccc}
            \toprule[1.5pt]
            \multirow{2}{*}{{\makecell[c]{Total Num.\\ of Image}}} &
            \multirow{2}{*}{{\small{\makecell[c]{Image\\ Classes}}}} &
            \multirow{2}{*}{{\small{\makecell[c]{Image\\per Class}}}} &
            \multicolumn{3}{c}{{Metrics(\%)}} \\
            & & & {Cls.Acc} & {Seg.F1} & {Seg.AUC} \\
            \midrule
            2,800 & 10 & 280 & 63.1 & 25.5 & 79.5 \\
            28,000~ & 100 & 280 & 74.8 & 32.5 & 79.9 \\
            180,000 & 100 & 1800 & 91.3 & 52.9 & 86.9 \\
            360,000~ & 200 & 1800 & 91.3 & 52.9 & 87.0 \\
            500,000~ & 500 & 1000 & 91.3 & 52.9 & 87.0 \\				
            \bottomrule[1.5pt]
        \end{tabular}}
        \captionof{table}{
Dataset Scale Experiment: Effect of Dataset Scale on the Performance of Base Models (SegFormer-b0, ResNet-50).}
        \label{tab:benchmark_scale}

    \end{minipage}%
    \hspace{0.35cm}
    \begin{minipage}[t]{0.48\textwidth}
        \centering
        \renewcommand{\arraystretch}{1.02} 
        {\fontsize{7pt}{8pt}\selectfont 
        \begin{tabular}{c|cc}
            \toprule[1.5pt]
            \multirow{2}{*}{Degradation} & \multicolumn{2}{c}{Metrics(\%)}               \\
                                         & Cls.Acc & Seg.F1 \\ 
            \midrule
            -                            & 91.3                   & 52.9           \\
            JPEG compression             & 83.1                   & 45.7           \\
            Gaussian blur                & 90.9                   & 40.1           \\
            Downsample                   & 88.3                   & 37.1           \\ 
            \bottomrule[1.5pt]
        \end{tabular}}
        \captionof{table}{Degradation Experiment: Impact of Degradation on Base Models (SegFormer-b0, ResNet-50).}
        \label{fig:benchmark_deg}

    \end{minipage}
\end{figure*}

\subsection{Post-processing of Degradation Method}\label{sec:post}
When uploaded to the Internet, images will encounter various post-processing. These transformations can pose challenges for image forensics methods. Table \ref{fig:benchmark_deg} investigates the impact of degradation. The baseline models are trained on the clean data and tested on the test set with a single degradation. Experimental results show that these degradations make the identification difficult. To reveal the real-world scenarios, random degradation  (JPEG compression, downsampling and Gaussian blur) is performed on the dataset.

\subsection{Benchmark Settings}\label{sec:setting}
\textbf{Benchmark Description:} 
GIM consists of four subsets. GIM-SD, GIM-GLIDE and GIM-DDNM which contain data from ImageNet manipulated by Stable Diffusion, GLIDE and DDNM, respectively. Additionally, there is a cross-distribution subset  GIM-VOC which contains data from VOC manipulated by Stable Diffusion. 

\noindent \textbf{Metrics:} 
We evaluate the performance of the IMDL method on both the image manipulation detection task and localization task. For the detection task, we use accuracy (Cls.acc) as our evaluation metric. For the localization task, the pixel-level AUC and F1 score are adopted. 

\noindent \textbf{Settings:} Two settings are proposed to evaluate the performance and generalization. In the mix-generator setting, the models are jointly trained on the GIM-SD, GIM-GLIDE and GIM-DDNM training set and tested on the correspondence test dataset respectively to evaluate the performance. In the cross-generator setting, the models are trained on the GIM-SD training set and tested on the GIM-GLIDE, GIM-DDNM and GIM-VOC test sets to explore the generalization.

\section{Method}
\begin{figure*}[!t]
	\centering
	\includegraphics[width=0.83\linewidth]{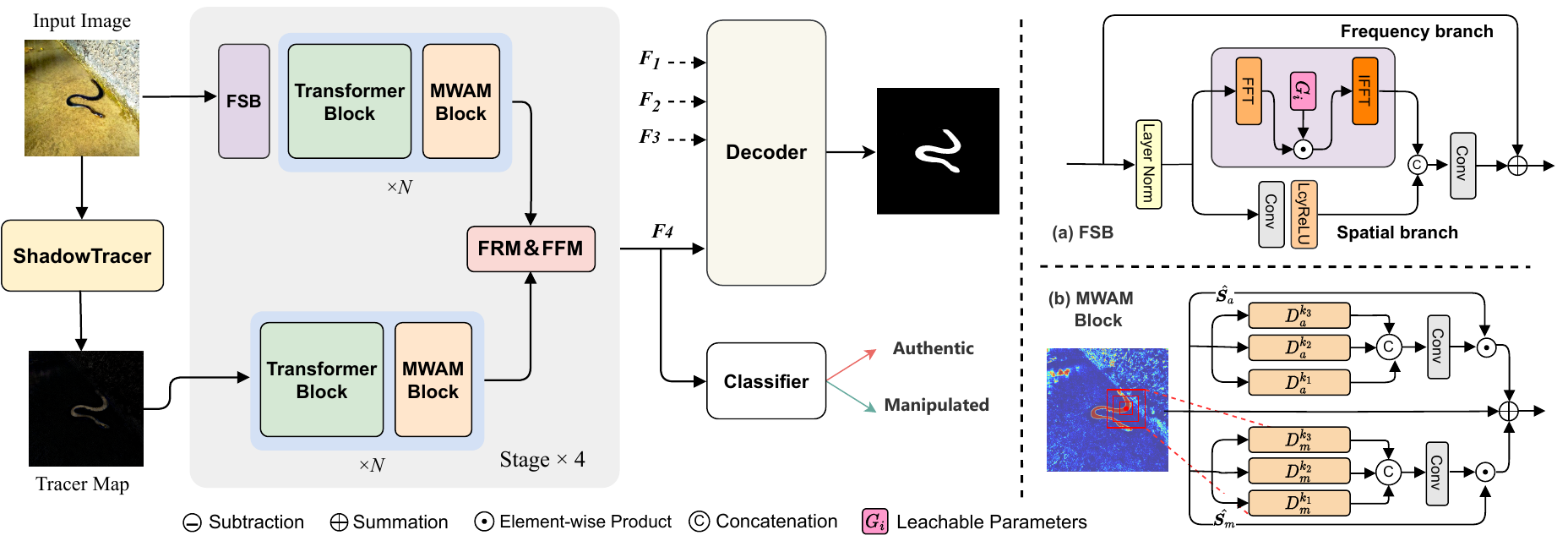}

	\caption{GIMFormer architecture. 
    ShadowTracer extracts trace map $t$ from the input image $x$. The encoder combines $x$ and $t$ to generate pyramid features $F_i$ across four stages, which are sent to the decoder for manipulation detection and localization.
    }
	\label{fig:pipe}

\end{figure*}

To address the challenges of generative manipulation, we propose GIMFormer that adopts a dual encoder and decoder architecture. Our framework includes several components: the ShadowTracer, the Frequency-Spatial Block (FSB), and the Multi Windowed Anomalous Modeling (MWAM) module. Figure \ref{fig:pipe} gives an overview of the framework. 
 For the RGB image input $x$, we first extract its learned trace map $t$. Then, both $x$ and $t$ are fed into a two-branch network, where the four-stage structure is used to extract pyramid features $F_i$  ($i\in [1,4]$). The RGB branch is composed of FSB, Transformer Block~\cite{xie2021segformer} and WMAM. The tracer branch consists of a Transformer Block and WMAM. In the fusion step, the feature rectification module (FRM) and feature fusion module (FFM)~\cite{zhang2023cmx} are used for feature fusion. The four-stage fused features are forwarded to the decoder for final detection $\hat{y}$ and location $\hat{M}$.

\subsection{ShadowTracer} \label{sec:1}

\begin{figure}[!t]
\centering
\includegraphics[width=1.0\linewidth]{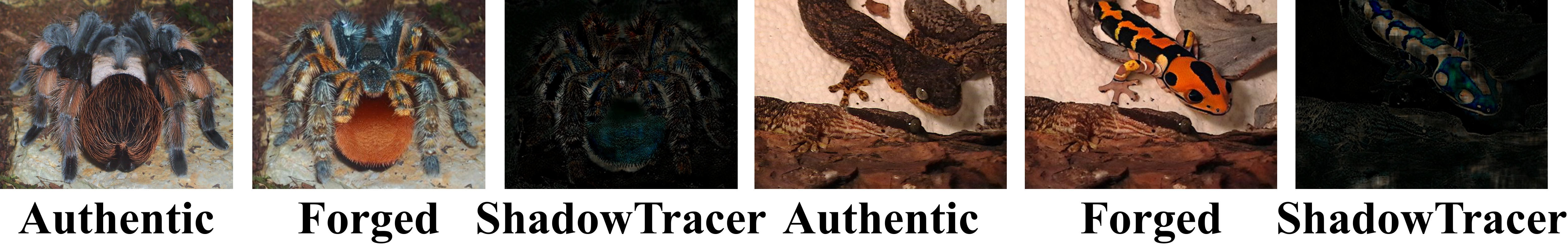}

\caption{ 
Generative manipulation leaves subtle traces, ShadowTracer identifies intrinsic patterns and reconstructs underlying tampering perturbations. 
}
\label{fig:shadow}
\end{figure}

Prior manipulation detection methods mainly focus on ``cheapfake'' and rely on visible traces. These artifacts include distortions and sudden changes caused by manipulation of the image structure. However, generative tampering makes significant alterations to the content with no apparent frequency or structural inconsistency. As shown in Figure \ref{fig:shadow}, these subtle traces are displayed with inherent patterns that are not visible traces with inconsistent edges.

ShadowTracer aims to capture the inherent characteristics and subtle traces of the generative models.
For a manipulated image, our objective is to learn a mapping $g_{\phi}$ to map the tampered image to its latent disturbed pixel values, where $g_{\phi}$ represents a neural network with trainable parameters $\phi$. Our key observation is that the differences introduced by generative models in data distribution exhibit inherent patterns, and deep neural networks can attempt to reconstruct these variations.
At the training stage, we generate pairs of the image $x_i$ and the tampered image $G(x_i) $, the manipulation trace can be calculated by $ t_i = G(x_i)-x_i$. The objective function for training $g_{\phi}$ can be formulated as :
\begin{equation}
	\min_{\boldsymbol{\phi}}\Big\{\mathcal{L}_r(g_{\boldsymbol{\phi}}\left(G\left(\mathbf{x}_i\right))\right),t_i)\Big\}
	\label{eq:2}
\end{equation}
where $\mathcal{L}_r(\mathbf{x},\mathbf{y})=\|\mathbf{x}-\mathbf{y}\|_2$. Furthermore, the mapping network needs to detect subtle tampering traces and be robust to various real-world image degradations.
For this reason,  image pairs are generated by mixing original and manipulated images and incorporating diverse degradation operations at the training stage.
Specifically, given an input image $I$,  we segment the region of interest and perform generative manipulation to obtain $I_m$.  The mix-up~\cite{zhang2017mixup} strategy is utilized to the original and manipulated images to obscure obvious tampering traces. Following this, we subject the images to the degradations mentioned above to obtain the final manipulated image. The network is trained on 64$\times$64 pixel patches randomly sampled from the dataset and the loss in Eq. \ref{eq:2} is adopted.

\subsection{Frequency-Spatial Block} \label{sec:2}
When degradation operations are applied, artifacts in manipulated images are tricky to perceive. To improve the local expressive and harvest discriminative cues in manipulated images, we design a Frequency-Spatial Block (FSB) to extract forgery features in the frequency and spatial domains.

Inspired by the recent work~\cite{rao2021global,lee2021fnet,zhang2022swinfir}, FSB consists of two branches: a frequency branch and a spatial branch as shown in Figure \ref{fig:pipe}.  In the frequency branch, the input $X$ is converted into the frequency domain $\mathcal{F}_T(X)$ using the  2D FFT. A learnable filter  $G_i$ is multiplied to modulate the spectrum and capture the frequency information. Subsequently, the inverse FFT is applied to convert the feature back to the spatial domain, resulting in the extraction of frequency-aware features $X_\text{f}$. In the spatial branch, the input $X$ is processed through convolution layers and LeakyReLU function to enhance the expressiveness of the features and obtain refined spatial features $X_\text{s}$. Then  $X_\text{f}$ and $X_\text{s}$  are concatenated and passed through convolution layers and the LeakyReLU function to obtain enhanced information, which is then combined with the original input $X$ through element-wise summation. The total process can be formulated by:

\begin{equation}
\begin{aligned}
	&X_\text{f}=\hat{\mathcal{F}}_T(\mathcal{F}_T(X)\odot G_i) \\
	&X_\text{s}=\mathrm{Conv_L}\left(\mathrm{Conv}\left(X\right)\right) \\
	&X_{out}=\mathrm{Conv_L}([X_{\mathrm{f}},X_{\mathrm{s}}])+X,
\end{aligned}
\end{equation}
where  $\odot$ denotes the Hadamard product , $\mathrm{Conv_L}$ denotes convolution with LeakyReLU and $\begin{bmatrix}\cdot\end{bmatrix}$  denotes concatenation.

\subsection{Multi Windowed Anomalous Modelling Module} \label{sec:3}
Image manipulation causes discrepancies at the pixel level. Genuine pixels are expected to exhibit consistency with neighboring pixels, while manipulated pixels may deviate and display anomalies.  Motivated by previous works~\cite{wu2019mantra,kong2023pixel} that explore local inconsistencies.  To effectively capture the pixel-level inconsistency between the manipulated and real region, we introduce the Multi Windowed Anomalous Modelling (MWAM)  module to model these differences at multiple scales for fine-grained features.

As shown in Figure \ref{fig:pipe}, for input feature ${F\in H\times W\times C}$, we calculate the difference between  pixel and its surroundings within a local window in two branches via Eq.\ref{eq:6}.

\begin{equation}
	\begin{aligned}
		D^k_u[i,j]=(F[i,j]-F^k_u[i,j])/\sigma^*,\\
		\sigma^*=\text{maximum}(\sigma(F),1e^{-5}+w_\sigma)
	\end{aligned}
	\label{eq:6}
\end{equation}
where $u\in \{a,m\}$ denotes average or maximum branches, $\sigma(F)$ is the standard deviation of $F$, and $w_\sigma$ is a learnable non-negative weight vector of the same length as $\sigma$, $F_{a}^{k}$ and $F_{m}^{k}$ are calculated from the average and maximum values of the $k\times k$ windows in each pixel. Different sizes $k$ are selected to model the inconsistency at different scales. Then, the obtained $N = 3$ different-scale $D^k_a$ and $D^k_m$ are concatenated and fed into a convolutional network to obtain an anomaly map $M_a$ and $M_m$ of the same size as the original input. 
 Additionally, the anomaly score mask $\hat{\boldsymbol{S_u}}\in H\times W$ of the feature is computed using.
\begin{equation}
	\begin{aligned}
		\hat{\boldsymbol{f}_u} & =\mathrm{DConv}\left(\boldsymbol{f}\right), \\
		\hat{\boldsymbol{S}_u} & =\operatorname{Sigmoid}\left(\mathrm{Conv}(C, 1)\left(\hat{\boldsymbol{f}_u}\right)\right),
	\end{aligned}
	\label{eq:3}
\end{equation}
where the $\mathrm{D}{\operatorname{Conv}}$ means a $3\times3$ Depth-Wise convolution layer.
The element-wise multiplication between anomaly score $\hat{\boldsymbol{S}_u}$ and the anomaly map $M_u$ capture the anomaly information. Next, we calculate the element-wise summation between the resulting anomaly-aware map and the input feature map $X$ to obtain an anomaly-sensitive feature map.  The whole process can be described as:
\begin{equation}
	\hat{X}=X+\hat{\boldsymbol{S}_a}\times M_a+\hat{\boldsymbol{S}_m}\times M_m
	\label{eq:8}
\end{equation}

\begin{table*}[!t]    
\fontsize{7}{8}\selectfont 

	\begin{center}
        \renewcommand{\arraystretch}{1.06}
        \setlength{\tabcolsep}{3pt} 

		\begin{tabular}{c|cc|ccc|ccc|ccc}
			\toprule
			\multirow{2}{*}{Method} & \multirow{2}{*}{Params} & \multirow{2}{*}{GFLOPS} & \multicolumn{3}{c|}{GIM-SD}            & \multicolumn{3}{c|}{GIM-GLIDE}         & \multicolumn{3}{c}{GIM-DDNM}          \\
			&                                  &                                 & {Cls.Acc} & {F1} & {AUC}   & {Cls.Acc} & {F1} & {AUC}   & {Cls.Acc} & {F1} & {AUC}  \\  \midrule
			ManTranet\cite{wu2019mantra}                        & \textbf{4.0}                            & 1009.7                            & 61.1                & 37.5       & 80.8          & 71.0                & 49.1       & 83.3          & 54.0                & 33.1       & 74.9         \\
			MVSS-Net\cite{dong2022mvss}                         & 146.9                            & 160.0                             & 56.1                & 23.2       & 72.0          & 61.3                & 33.1       & 74.9          & 49.2                & 14.1       & 70.1         \\
			SPAN\cite{hu2020span}                             & 15.4                                 & \textbf{30.9}                                & 53.2                & 35.6       & 79.3          & 60.0                & 39.5       & 81.2          & 59.2                & 32.1        & 73.6         \\
			PSCC-Net\cite{liu2022pscc}                             & 4.1                              & 107.3                                & 52.3            & 31.5       & 83.8          & 66.5            & 53.7       & 86.4          & 56.3            & 41.8       & 85.8          \\
			ObjectFormer$\ddagger$\textsuperscript{\ref{fn:code}}\cite{wang2022objectformer}                     &14.6                                 &  249.6                             &  59.1                &26.8             & 85.2               & 70.1                 &40.1             & 85.2               & 54.3                & 33.1            &  86.8             \\
			Trufor$\ddagger$\textsuperscript{\ref{fn:code}}\cite{guillaro2023trufor}                           & 67.8                                 &90.1                                 & 67.1                 &44.1             & 84.5               & 80.2                 &  59.3           & 93.0               & 63.3                 &  44.5           &  87.6             \\  
			        IML-VIT\cite{ma2024imlvitbenchmarkingimagemanipulation} &91.0 &136.0 &65.2 &53.9 &84.3	&71.3 &69.3 &92.1	&71.2 &53.9 &85.7\\
			SegFormer\cite{xie2021segformer}                       &  27.5                                & 41.3                               & 64.3            & 46.2       & 83.3        & 78.1            & 56.8       & 88.7        & 69.3            & 40.2       & 84.6         \\

			\midrule
			
			\textbf{GIMFormer (Ours)}                             & 95.9                             & 96.2                            & \textbf{70.9}            &\textbf{ 58.6}       & \textbf{88.2} & \textbf{83.9}            & \textbf{77.3}      & \textbf{95.42} & \textbf{76.7}            & \textbf{56.3}      &\textbf{88.3} \\ \bottomrule
		\end{tabular}
        		\caption{
Benchmarking IMDL models to evaluate performance.
   Cls.Acc(\%), AUC(\%) and F1(\%) Params(M) are reported. 
		}
        \label{tab:benchmark_performance}

	\end{center}

\end{table*}

\begin{table*}[!t] \fontsize{7}{8}\selectfont 
        \centering
	\begin{center}
        \renewcommand{\arraystretch}{1.08}
        \setlength{\tabcolsep}{2pt} 
		\begin{tabular}{c|ccc|ccc|ccc|ccc}
			\toprule
			\multirow{2}{*}{Method} & \multicolumn{3}{c|}{GIM-SD}            & \multicolumn{3}{c|}{GIM-VOC}       & \multicolumn{3}{c|}{GIM-GLIDE}         & \multicolumn{3}{c}{GIM-DDNM}                      \\
			& Cls.Acc & F1 & AUC   & Cls.Acc & F1 & AUC   & Cls.Acc & F1 & AUC   & Cls.Acc & F1              & AUC \\ 
			\midrule
			ManTranet\cite{wu2019mantra}                        & 73.1               & 43.2       & 80.2          & 63.2                & 27.4       & 72.7          & 58.2                 & 24.5       & 74.6          & 39.5                 & 16.8                     & 58.3        \\
			MVSS-Net\cite{dong2022mvss}                         & 56.1                & 25.1       & 82.2          & 56.3                & 21.2        & 73.2          &53.2                 & 23.17       & 73.1          & 48.1                 & 10.1                         &50.2              \\
			SPAN\cite{hu2020span}                             & 52.2                & 47.3       & 56.9          & 52.2                 & 29.9       & 55.2          & 50.0                & 30.1       & 56.1          & 44.2                & 14.3 & 56.7        \\
			PSCC-Net\cite{liu2022pscc}                             & 58.2            & 48.4       & 87.3          & 54.8            & 39.4       & 83.8           & 51.1            & 29.6        & 79.8          & 40.5            & 4.4                      & 48.5        \\
			objectFormer$\ddagger$\textsuperscript{\ref{fn:code}}\cite{wang2022objectformer}                     &67.1                  & 39.5            & 87.9               & 58.2                 & 29.1            & 81.2               & 54.2                & 17.7            &  81.0              & 49.2                 & 7.2                         & 59.0             \\
			Trufor$\ddagger$\textsuperscript{\ref{fn:code}}\cite{guillaro2023trufor}                           & 74.0                 & 49.9            & 86.2               & 65.2                 &31.9             & 80.1               & 50.1                 & 22.4            & 76.3               & 49.2                 &  5.7                        & 53.1             \\
            IML-VIT\cite{ma2024imlvitbenchmarkingimagemanipulation} 
            &77.1 &58.1 &88.9	&54.3 &47.2 &83.2	&52.1 &23.1 &71.1	&50.3 &7.2 &52.3 \\
			SegFormer\cite{xie2021segformer}                        & 71.9                & 53.1       & 87.1         & 60.0                & 28.2         &81.0                & 54.1                & 21.0           & 75.1              & 50.2                 &4.7                          & 51.3             \\
			\midrule
			\textbf{GIMFormer (Ours)}                             & \textbf{78.9}            & \textbf{61.7}       & \textbf{90.6} & \textbf{67.2}             & \textbf{50.0}       & \textbf{84.0} & \textbf{67.0}            & \textbf{39.3}       & \textbf{81.0} & \textbf{52.3}             & \textbf{19.4}                   & \textbf{60.1}         \\ \bottomrule
		\end{tabular}
	\end{center}
    	\caption{Benchmarking IMDL models to evaluate generalization. 
        Cls.Acc(\%), AUC(\%) and F1(\%) are reported.
		}
        \label{tab:benchmark_generalization}


\end{table*}

\subsection{Loss Function}
For detection, we adopt a lightweight backbone in ~\cite{wang2020deep} on the fourth stage feature for binary prediction $\hat{y}$. For localization, we utilize the MLP decoder ~\cite{xie2021segformer} as the segmentation head to obtain a predicted mask $\hat{M}$. 
Given the ground-truth label $y$ and mask $M$, we train GIMFormer with the following
objective function: 
\begin{equation}
	{\mathcal{L}}={\mathcal{L}}_{c l s}(y,{\hat{y}})+{\mathcal{L}}_{s e g}(M,{\hat{M}}),
	\label{eq:9}
\end{equation}
where both ${\mathcal{L}}_{c l s}$ and ${\mathcal{L}}_{s e g}$ are binary cross-entropy loss.

\subsection{Implementation Details}
Our approach includes two separate training steps. First, we train the ShadowTracer using the dataset generated from  ImageNet. This training process follows a similar data generation method as mentioned in the former chapter. Then, we train the encoder and decoder of the model according to the two settings in GIM, as described in the previous section. We train our models on 8 V100 GPUs with an initial learning rate of $6e^{-5}$
which is scheduled by the poly strategy with power 0.9 over 20 epochs.
The optimizer is AdamW ~\cite{loshchilov2017decoupled} with epsilon $1e^{-8}$ weight decay $1e^{-2}$, and the batch size is 4 on each GPU.

\subsection{Comparison with state-of-the-art methods}
We compare our methods with various state-of-the-art IMDL methods \footnote{\label{fn:code} $\ddagger$ indicates that the original paper does not provide the code, we reproduce the code and evaluate it under the same settings.} and the vanilla SegFormer (MiT-B2)~\cite{xie2021segformer}. As shown in Table \ref{tab:benchmark_performance} and Table \ref{tab:benchmark_generalization},  these methods are tested in two settings to evaluate the performance. Note that some methods are not explicitly designed for image-level detection, in which case we use the maximum of the prediction map as the detection statistic. All methods are immersed in the same implementation details. \ More experiments are included in the Appendix.

\begin{figure*}[!t]
\centering
\includegraphics[width=0.73\linewidth]{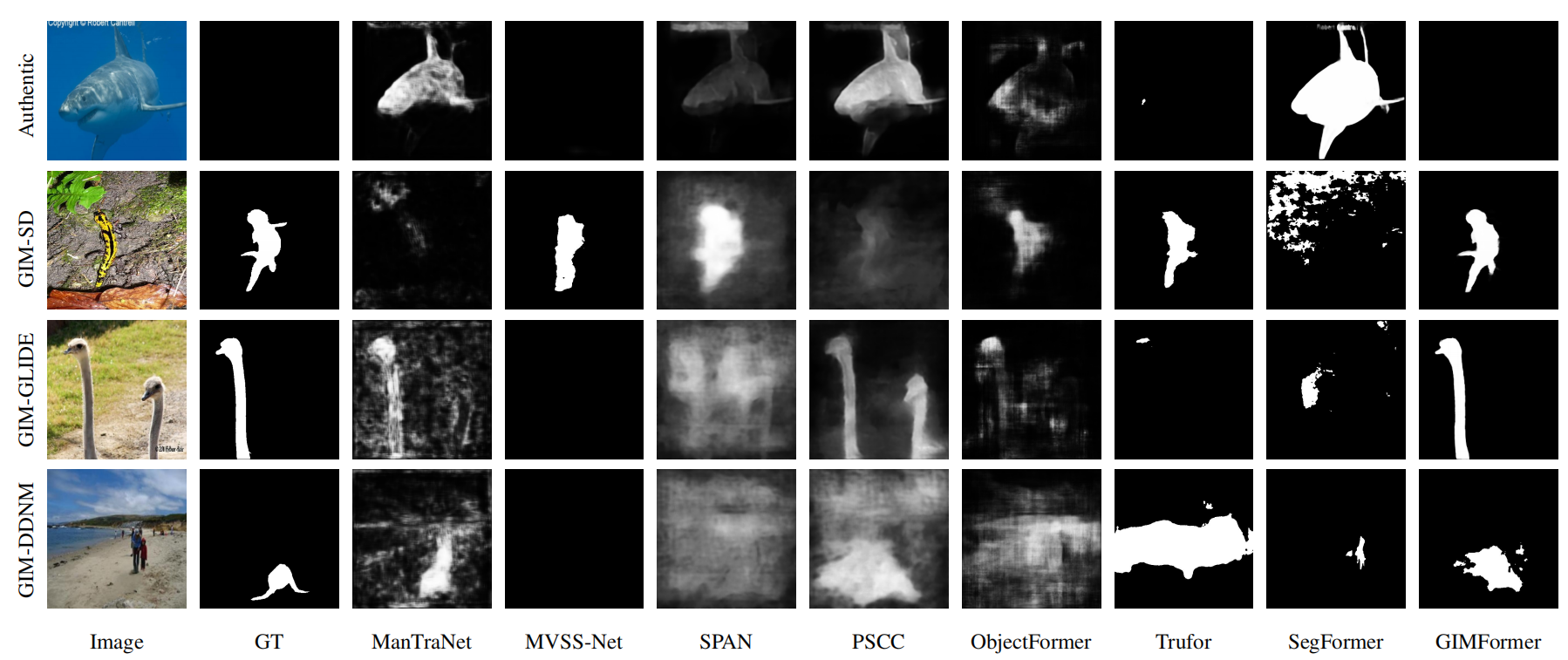}

\caption{ Qualitative results on GIM of comparing GIMFormer with state-of-the-art methods performance.
}
\label{fig:vis}
\end{figure*}

\noindent \textbf{Mix-generator comparison.}
Table \ref{tab:benchmark_performance} reports the performance of all methods.  GIMFormer outperforms other methods, demonstrating its superior ability to identify generative manipulation. Existing methods exhibit low pixel-level F1 scores on this benchmark, indicating that tampering areas are not accurately identified. The qualitative results for visual comparisons are illustrated in Figure \ref{fig:vis}.

\begin{figure*}[!t]
\centering
\includegraphics[width=0.7\linewidth]{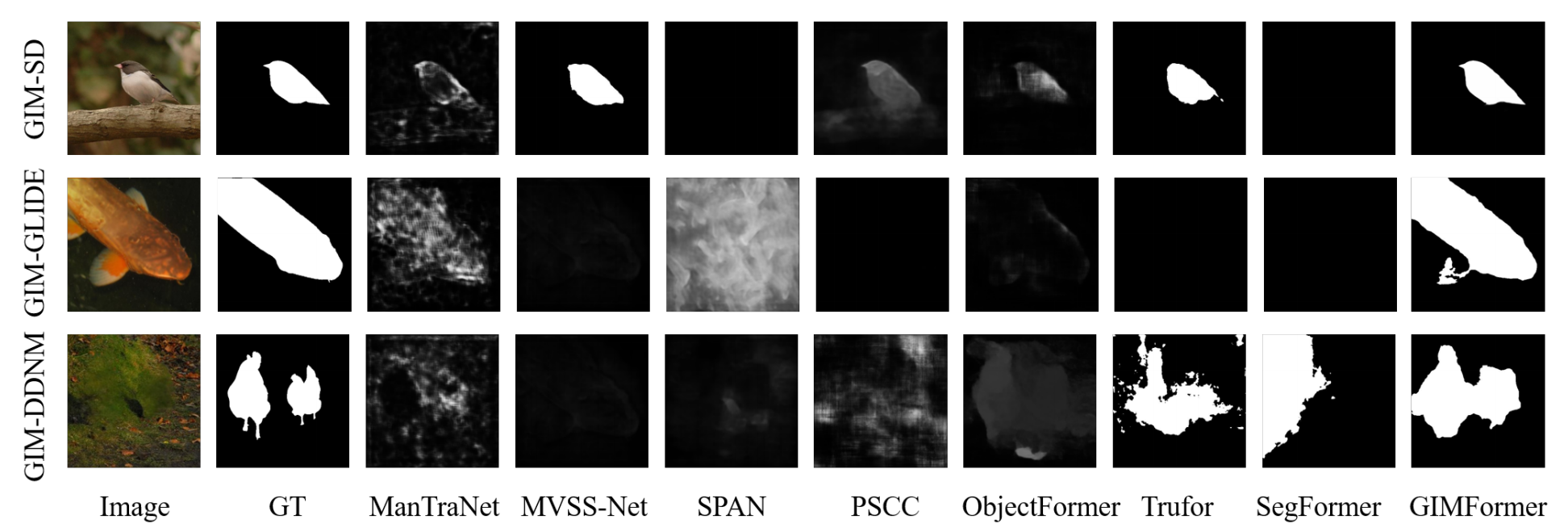}
\caption{ Qualitative results on GIM of comparing GIMFormer with state-of-the-art methods generalization capability. 
}
\label{fig:vis_gen}
\end{figure*}

\noindent \textbf{Cross-generator comparison.}
Table \ref{tab:benchmark_generalization} reports the generalization of all the methods. The results show that  GIMFormer outperforms all other methods in both in-domain and cross-domain. 
For in-domain experiments, our method catches the subtle artifacts inherent in manipulation and accurately localizes them. Other methods may encounter confusion as they attempt to learn specific content, potentially leading to challenges in accurately detecting and localizing generative tampering patterns.
For cross-domain experiments, GIMFormer demonstrates well generalization in detecting manipulation using different generative models, as shown in the results of the GIM-GLIDE and GIM-DDNM testsets. Besides GIMFormer works well on data generated by the same generator from different distributions, while existing methods have an obvious performance drop, as shown in the results of the GIM-VOC testset. The qualitative results for visual comparisons are illustrated in Figure \ref{fig:vis_gen}.

\subsection{Ablation Analysis}
\noindent \textbf{Effectiveness of the proposed module.}
 We consider a simple baseline proposed in ~\cite{xie2021segformer} and gradually integrate new key components. Experiments are carried out on the GIM-GLIDE testset. The quantitative results are listed in Table \ref{table:ablation}. The result shows that ShadowTracer brings significant improvements to the vanilla baseline. With the MWAM, there is an increase of 6.29\% in F1 and 2\% in AUC, which indicates that the differential information at multiple scales is crucial for accurate tampering localization.  The use of FSB to dynamically harvest complementary frequency and spatial cues improves performance, particularly in detection.
The results verify that ShadowTracer, FSB and MWAM effectively improve the performance of the baseline model. 

\begin{table}[!t]
    \fontsize{7}{8}\selectfont
    \centering
    \renewcommand\tabcolsep{2pt}
    \renewcommand{\arraystretch}{1.05}
    \begin{tabular}{l|ccc}
      \toprule[1.2pt]
      \multirow{1}{*}{Variants} & Cls.Acc & F1 & AUC \\ 
      \midrule
      Baseline                  & 78.1   & 56.8 & 88.7 \\
      +ShadowTracer                        & 79.9   & 67.1 & 92.0 \\
      +ShadowTracer+WMAM                   & 80.2   & 73.4 & 94.0 \\
      +ShadowTracer+WMAM+FSB               & \textbf{83.9} & \textbf{77.3} & \textbf{95.4} \\
      \bottomrule[1.2pt]
    \end{tabular}
    \caption{Ablation results of GIMFormer variants. Cls.Acc(\%), AUC(\%) and F1(\%) scores are reported.}
    \label{table:ablation}

\end{table}

\begin{table}[!t]
        \fontsize{7}{8}\selectfont 
        \centering
        \renewcommand{\arraystretch}{1.05}
        \setlength{\tabcolsep}{2pt} 

      \begin{tabular}{c|cccccc}
        \toprule[1.5pt]
        \multirow{2}{*}{Variants} & \multicolumn{3}{c}{GIM-GLIDE} & \multicolumn{3}{c}{GIM-DDNM} \\
        & Cls.Acc & F1 & AUC & Cls.Acc & F1 & AUC \\ \midrule
        GIMFormer w ShadowTracer & 83.1 & 78.1 & 95.4 & 77.4 & 58.8 & 89.4 \\
        GIMFormer w/o ShadowTracer & 80.0 & 68.9 & 91.3 & 73.1 & 51.3 & 86.1 \\
        \bottomrule[1.5pt]
      \end{tabular}
        \caption{Generalization Experiments of ShadowTracer. 
        Cls.Acc(\%), AUC(\%) and F1(\%) scores are reported.}
        \label{ST}

\end{table}

\noindent \textbf{Generalization of ShadowTracer.}  To evaluate the generalization of ShadowTracer, we train ShadowTracer using data generated by Stable Diffusion and subsequently hold its weights fixed. We proceed to train the backbone on the GIM-GLIDE and GIM-DDNM trainsets, with and without the incorporation of ShadowTracer. The result in Table \ref{ST} reveals that leveraging the pretrained ShadowTracer increases performance in cross-generator IMDL tasks. 
Across GIM-GLIDE and GIM-DDNM, ShadowTracer achieves up to 9\% F1, 4\% AUC and 4\% accuracy improvements.

\section{Conclusion}

We address the challenge of detecting and locating generative manipulation and provide a reliable database GIM for AIGC security. This dataset leverages multiple generators to provide diverse generative manipulation data. Based on this, we design a benchmark for IMDL methods with two settings. We also introduce GIMFormer, a novel transformer-based IMDL framework. Extensive experiments demonstrate that GIMFormer achieves SOTA performance.

\appendix

\section*{Appendix}
In this Appendix, we include more details of our work:  
 (1) The details of the proposed GIM Datset and compare with related work  (2) Details of the GIMFormer architecture and implementation 
 (3) The additional ablation analysis of GIMFormer
 (4) More Visualizations on GIM dataset. (5)The societal impact of our work . (6) The ethics statement of our work.

\section{GIM Dataset}
\label{sec:rationale}

\subsection{GIM Dataset Configuration}
The specific quantities and details of the dataset are presented in Table  \ref{table:config}, GIM has a total number of 1,140k generative manipulated images with their corresponding origin images.  Moreover, the dataset can be split into subsets.
GIM-SD-all utilizes the training set from Imagenet-1k~\cite{deng2009imagenet} and employs the Stable Diffusion generator for image manipulation dedicated to manipulation research.  Through the analysis of the data scale, GIM-SD, GIM-GLIDE and GIM-DDNM individually select a random set of 100 classes from the Imagenet-1k training dataset and the entire test data to generate their respective training and test sets to benchmark the image manipulation detection and location (IMDL)methods. GIM-VOC is a cross-data-distribution test set constructed using the test set from the PASCAL VOC~\cite{everingham2010pascal} dataset and Stable Diffusion techniques.

\begin{table*}[!t]
    \centering
    \begin{minipage}[t]{.45\linewidth}
        \centering
        \fontsize{12}{14}\selectfont
        \caption{Parameters for GIM degradation. Three degradation methods are employed with variable parameters to emulate real-world scenarios.}
            \vspace{0.7mm}

            \resizebox{0.75\textwidth}{!}{
        
        \begin{tabular}{cc}
            \toprule[1.5pt]
            Post-processing        & Parameter \\
            \hline
            JPEG compression        & 75, 80, 90 \\
            Gaussian blur            & 3, 5      \\
            Downsampling             & 0.5, 0.67 \\
            \bottomrule[1.5pt]
        \end{tabular}}
        \label{table:parma}
    \end{minipage}%
        \hspace{0.1cm}
    \begin{minipage}[t]{.50\linewidth}
        \centering
        \caption{Basic configuration about the GIM Dataset.  GIM has five subsets, leveraging various generators and data sources.}
        \resizebox{0.85\textwidth}{!}{
\begin{tabular}{cccc}
	\toprule[1.2pt]
	GIM         & \begin{tabular}[c]{@{}c@{}}Tampreing\\ Type\end{tabular} & \begin{tabular}[c]{@{}c@{}}Number of \\ Trainset\end{tabular} & \begin{tabular}[c]{@{}c@{}}Number of \\ Testset\end{tabular} \\ \hline
	GIM-SD-all  & reprint                                                  & 1890K                                                          & -                                                             \\
	GIM-SD      & reprint                                                  & 180K                                                           & 70K                                                           \\
	GIM-GLIDE   & reprint                                                  & 140K                                                           & 70K                                                           \\
	GIM-DDNM    & removal                                                  & 125K                                                           & 50K                                                           \\
	GIM-VOC& reprint                                                  & -                                                              & 4.5K                                                          \\ \bottomrule[1.2pt]
\end{tabular}}
        \label{table:config}
    \end{minipage}
\end{table*}

To simulate real-world scenarios and comprehensively evaluate the IMDL methods, GIM incorporates three degradation methods (JPEG compression, Gaussian blur, and downsampling). Table \ref{table:parma} presents the parameters for each degradation group, which are randomly chosen.

The GIM data format consists of JPG images paired with PNG labels. Filenames ending with "\_f" indicate tampered images. The masks segmented by SAM~\cite{kirillov2023segment} serve as ground truth for manipulation detection, containing two labels denoted by 0 and 255, representing the original and manipulated categories, respectively.

\subsection{Dataset Comparison.}
GIM focuses on generative local manipulation in natural images and provides manipulated region labels,  specifically for image manipulation detection and localization tasks. This distinguishes GIM from the existing datasets focused on faces or limited to detection (classification). As shown in Table \ref{table:datacom}, we compare several local image manipulation datasets. The existing image forensic datasets are primarily constructed using traditional manipulation techniques. These datasets commonly employ manual or random manipulation to generate data. AutoSplice~\cite{jia2023autosplice}, HiFi-IFDL\cite{guo2023hierarchical}, CoCoGlide ~\cite{guillaro2023trufor} dataset utilizes generative algorithms for image manipulation. However, they rely on one type of manipulation and suffer from a small data volume.

In contrast, GIM achieves image manipulation by employing generative methods for localized image alteration. It encompasses a variety of generators and manipulation methods. Simultaneously, it stands as a large-scale comprehensive dataset for the IMDL community.

\begin{table*}[!t]
	\caption{Summary of previous image manipulation datasets and GIM. We showcase the number of data entries within each dataset and the manipulation techniques they encompass. *denotes that HiFi-IFDL includes entirely synthesized images}
	\centering
\label{table:datacom}
\resizebox{\textwidth}{!}{
\begin{tabular}{cccccccc}
\hline
\multirow{2}{*}{Dataset}               & \multicolumn{2}{c}{Num. Images}     & \multicolumn{3}{c}{Handcraft Manipulations} & \multicolumn{2}{c}{AIGC Manipulations}    \\
                                       & \# Authetic Image & \# Forged Image & Splicing      & Copy-move    & Removal      & Removal              & Generation         \\ \hline
Columbia Gray~\cite{ng2004data}        & 933               & 912             & \ding{51}     & \ding{55}    & \ding{51}    & \ding{55}            & \ding{55}          \\
Columbia Color~\cite{hsu2006detecting} & 183               & 180             & \ding{51}     & \ding{55}    & \ding{51}    & \ding{55}            & \ding{55}          \\
MICC-F2000~\cite{amerini2011sift}      & 1,300             & 700             & \ding{51}     & \ding{55}    & \ding{51}    & \ding{55}            & \ding{55}          \\
VIPP Synth. ~\cite{amerini2011sift}    & 4,800             & 4,800           & \ding{51}     & \ding{55}    & \ding{51}    & \ding{55}            & \ding{55}          \\
CASIA V1.0 ~\cite{dong2013casia}       & 800               & 921             & \ding{51}     & \ding{51}    & \ding{51}    & \ding{55}            & \ding{55}          \\
CASIA V2.0 ~\cite{dong2013casia}       & 7,200             & 5,123           & \ding{51}     & \ding{51}    & \ding{51}    & \ding{55}            & \ding{55}          \\
Wild Web~\cite{zampoglou2015detecting} & 90                & 9,657           & \ding{51}     & \ding{51}    & \ding{51}    & \ding{55}            & \ding{55}          \\
NC2016 ~\cite{guan2019mfc}             & 560               & 564             & \ding{51}     & \ding{51}    & \ding{51}    & \ding{55}            & \ding{55}          \\
NC2017  ~\cite{guan2019mfc}            & 2,667             & 1,410           & \ding{51}     & \ding{51}    & \ding{51}    & \ding{55}            & \ding{55}          \\
MFC2018 ~\cite{guan2019mfc}            & 14,156            & 3,265           & \ding{51}     & \ding{51}    & \ding{51}    & \ding{55}            & \ding{55}          \\
MFC2019 ~\cite{guan2019mfc}            & 10,279            & 5,750           & \ding{51}     & \ding{51}    & \ding{51}    & \ding{55}            & \ding{55}          \\
PS-Battles~\cite{heller2018ps}         & 11,142            & 102,028         & \ding{51}     & \ding{51}    & \ding{51}    & \ding{55}            & \ding{55}          \\
DEFACTO ~\cite{mahfoudi2019defacto}    & 0                 & 229,000         & \ding{51}     & \ding{51}    & \ding{51}    & \ding{55}            & \ding{55}          \\
IMD2020~\cite{novozamsky2020imd2020}   & 35,000            & 35,000          & \ding{51}     & \ding{51}    & \ding{51}    & \ding{55}            & \ding{55}          \\
SP COCO ~\cite{kwon2022learning}       & 0                 & 200,000         & \ding{51}     & \ding{55}    & \ding{55}    & \ding{55}            & \ding{55}          \\
CM COCO   ~\cite{kwon2022learning}     & 0                 & 200,000         & \ding{51}     & \ding{55}    & \ding{55}    & \ding{55}            & \ding{55}          \\
CM RAISE  ~\cite{kwon2022learning}     & 0                 & 200,000         & \ding{51}     & \ding{51}    & \ding{55}    & \ding{55}            & \ding{55}          \\
HiFi-IFDL\cite{guo2023hierarchical}    & -                 & 1,000,00*       & \ding{55}     & \ding{51}    & \ding{55}    & \ding{51}            & \ding{55}          \\
CoCoGlide~\cite{guillaro2023trufor}    & 0                 & 512             & \ding{55}     & \ding{55}    & \ding{55}    & \ding{51}            & \ding{55}          \\
AutoSplice  ~\cite{jia2023autosplice}  & 3,621             & 2,273           & \ding{55}     & \ding{55}    & \ding{55}    & \ding{55}            & \ding{51}          \\
GIM                                    & 1,140,000         & 1,140,000       & \ding{55}     & \ding{55}    & \ding{55}    & \textbf{ \ding{51} } & \textbf{\ding{51}} \\ \hline
\end{tabular}}
\end{table*}

\section{Implementation Details}\label{sec:imple}
\subsection{Architecture}
The GIMFormer backbone has an encoder-decoder architecture. The encoder is a dual-branch and four-stage encoder. For the input RGB image, ShadowTracer extracts its learned manipulation trace map of the same resolution as the image. The ShadowTracer adopts the architecture~\cite{zhang2017beyond} with 15 trainable layers, 3 input channels, 1 output channel.
Then, both the RGB and the trace map are fed into the parallel network, where the four-stage structure is employed to extract pyramidal features $F_i$  ($i\in [1,4]$). At each stage the RGB branch is first processed by the Frequency-Spatial Block (FSB), then two branches are gradually processed by Transformer Blocks ~\cite{xie2021segformer} and Multi Windowed Anomalous Modelling module(MWAM). 
The learnable parameters within FSB (Feature Synthesis Block)maintain an identical resolution as the input features. 
The window sizes within the Multi-Window Anomalous Modeling (MWAM)at each stage are as follows: \{3, 7, 9\}, \{7, 11, 15\}, \{9, 17, 25\}, and \{11, 21, 31\}.
The Transformer blocks are based on the Mix Transformer encoder B2 (MiT-B2)proposed for semantic segmentation and are pretrained on ImageNet. 
The Mix Transformer encoder uses self-attention and channel-wise operations, prioritizing spatial convolutions over positional encodings. To integrate information from the two branches, the cross-modal Feature Rectification Module (FRM) and Feature Fusion Module (FFM)~\cite{zhang2023cmx} are utilized. 
For location, we employ the All-MLP decoder proposed in~\cite{xie2021segformer}, which is a lightweight architecture formed by only 1$\times$1 convolution layers and bilinear up-samplers. For detection, we adopt the tail-end section of the light-weight backbone in ~\cite{wang2020deep}. It applies convolutional, batch normalization, and activation layers to extract features from input data. These extracted features are then pooled and passed through fully connected layers to generate the classification predictions.

\subsection{Training Process}
We conduct our experiments with PyTorch 1.7.0. All models are trained on a node with 8 or 4 V100 GPUs.

For the training of the ShadowTracer network $g_{\phi}$, We randomly selected 20,000 authentic images from ImageNet and correspondingly generated manipulated images. During comprehensive performance verification, all three generators (Stable Diffusion~\cite{rombach2022high}, GLIDE~\cite{nichol2021glide}, DDNM~\cite{wang2022zero})are used, whereas for generalization verification, only Stable Diffusion is employed. Throughout the training, the alpha blending parameter ranges between 0.5 and 1.0, randomly producing blended images subjected to three degradation types (JPEG compression, downsampling, Gaussian blur). The network is trained on 40$\times$40 pixel patches randomly sampled from the dataset. Training is conducted for roughly 300,000 iterations with a batch size of 64. An Adam~\cite{kingma2014adam} optimizer is employed, initialized with a learning rate of 0.001.

For the training of the GIMFormer backbone network. The input image is cropped to 512 $\times$ 512 during training.
We train our models for 40 epochs on the GIM dataset. The batch size is 4 on each of the GPUs.
The images are augmented by random resize with a ratio of 0.5–2.0, random horizontal flipping.

\section{ Additional Ablation Analysis}\label{sec:abl}

\noindent \textbf{The robustness of GIMFormer.}
To evaluate the robustness of GIMFormer, we sample 5000 images from the original three generator test sets without degradations.  The model is trained on the mixed-generator dataset. The model is trained on the mixed-generator dataset. The experimental results are presented in Table \ref{robust}, GIMFormer demonstrates robustness against various distortion techniques. This resilience underscores its ability to maintain stable performance in the face of various challenges posed by image distortions.

\begin{table*}[!t]
 	\caption{Robustness experiment of pixel-level Manipulation Localization F1(\%)
with various distortions}
\centering
\label{robust}
\resizebox{0.95\textwidth}{!}{
\begin{tabular}{c|cccccccc}
\toprule[1.5pt]
          & \multicolumn{1}{c}{\begin{tabular}[c]{@{}c@{}}No Dis-\\ tortion\end{tabular}} & \begin{tabular}[c]{@{}c@{}}Cmp\\ (q=90)\end{tabular} & \begin{tabular}[c]{@{}c@{}}Cmp\\ (q=80)\end{tabular} & \multicolumn{1}{c}{\begin{tabular}[c]{@{}c@{}}Cmp\\ (q=75)\end{tabular}} & \begin{tabular}[c]{@{}c@{}}Blur\\ (k=3)\end{tabular} & \begin{tabular}[c]{@{}c@{}}Blur\\ (k=5)\end{tabular} & \multicolumn{1}{c}{\begin{tabular}[c]{@{}c@{}}Downsample\\ (0.66X)\end{tabular}} & \multicolumn{1}{c}{\begin{tabular}[c]{@{}c@{}}Downsample\\ (0.5X)\end{tabular}} \\ \midrule
GIMFormer &61.7                                                                               &60.3                                                      &60.1                                                      &59.9                                                                          &58.7                                                          &58.7                                                          &61.0                                                                              &60.8                                                                              \\ \bottomrule[1.5pt]\end{tabular}}

\end{table*}

\noindent \textbf{The effect of the number of windows (N).}
We conduct a set of ablation experiments to study the performance of the MWAM Module.  To ensure fair comparisons, all experiments differ from each other only in the Windows setting. Experiments are carried out on the GIM-SD trainset and testset. As shown in Table \ref{table:add_ablation}, there is an overall incremental trend in the tampering location performance as the number of windows increases, while the impact of size variations remains relatively minor. For the sake of efficiency, we stop analyzing more windows. 
The results indicate that a favorable balance between accuracy and efficiency is achieved when $N$=3.

\begin{table}[!t]\fontsize{10}{12}
	\centering
	\caption{Ablation results on GIM-SD test dataset using different variants of GIMFormer, All detection Cls.Acc(\%) and localization AUC(\%) and F1(\%) scores are reported. Where {$D_{a/m}^{k}$} represents the number and dimensions of windows utilized in either the average or maximum branches. We investigate the window counts $N$ of 0, 1, 2,3 as well as the impact of only one branch on the module. In this table, the window sizes of the first layer are used to annotate, with subsequent layers decreasing in size.}
	\resizebox{0.48\textwidth}{!}{
         \renewcommand{\arraystretch}{1.2}  
		\begin{tabular}{r@{\hspace{8pt}}|@{\hspace{8pt}}c@{\hspace{8pt}}c@{\hspace{8pt}}c}
			
			\toprule[1.2pt]
			MWAM Variants                            & Cls.Acc &F1  & AUC \\ \midrule
			w\textbackslash{}o windows                                &72.13                          &58.12                      &88.81                      \\
			\{$D_{a\&k}^{11}$\} &74.33                          &58.34                      &89.65                      \\
			\{$D_{a\&k}^{11},D_{a\&k}^{21}$\} &                          76.61&                      60.03&                   89.92   \\
			\{$D_{a\&k}^{11},D_{a\&k}^{21},D_{a\&k}^{31}$\} &                         \textbf{78.96} &                       \textbf{61.75}&                      90.61 \\
			\{$D_{a\&k}^{17},D_{a\&k}^{29},D_{a\&k}^{41}$\} & 77.33    & 61.17 & 90.11 \\
			\{$D_a^{11},D_a^{21},D_a^{31}$\}                            &77.95      &60.77  &90.03  \\
			\{$D_m^{11},D_m^{21},D_m^{31}$\}                            &77.83      & 59.13 & 89.53 \\
			\{$D_{a\&k}^{11},D_{a\&k}^{21},D_{a\&k}^{31},D_{a\&k}^{41}$\} &78.11      &61.15  &\textbf{90.71}  \\ 
			\bottomrule[1.2pt]
	\end{tabular}}
	\label{table:add_ablation}
	
\end{table}

\section{More Visualizations on GIM}\label{sec:vis}
\begin{figure*}[!t]
	\centering
	\includegraphics[width=1.0\linewidth]{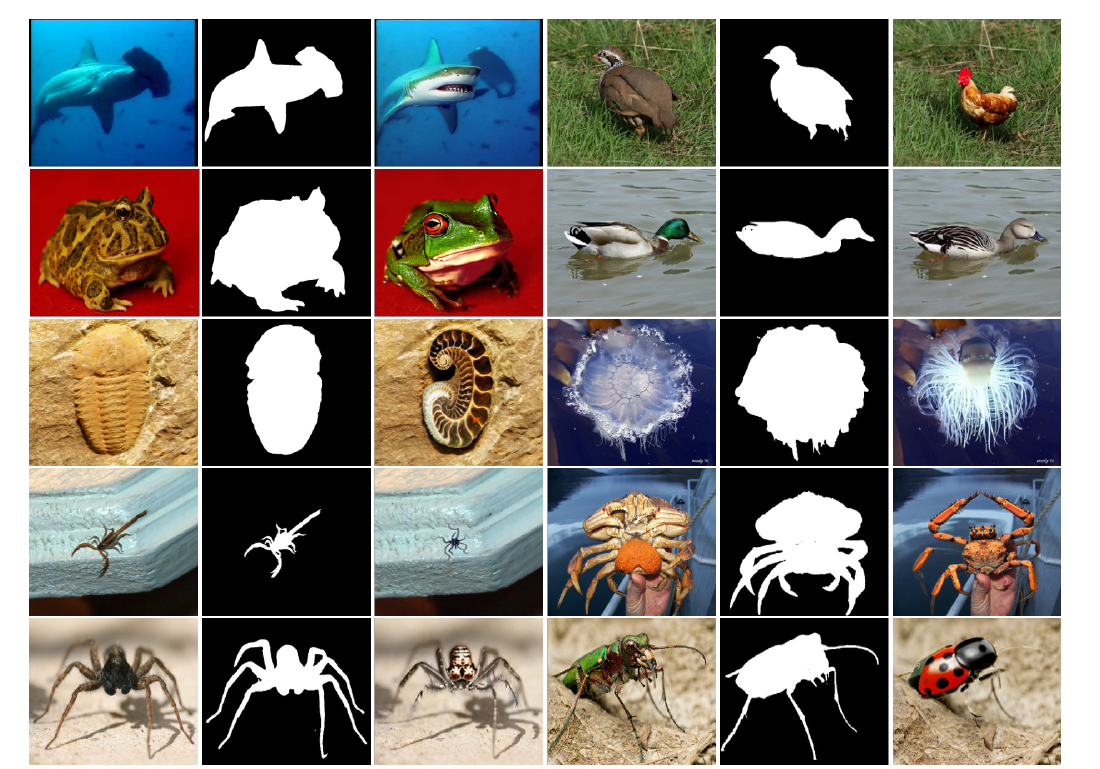}
	\caption{ Additional visualizations of GIM-SD. From left to right, the sequence is as follows: original image, tampered area, generative manipulated image, original image, tampered area, generative manipulated image.}
	\label{fig:SD}
\end{figure*}

\begin{figure*}[!t]
	\centering
	\includegraphics[width=1.0\linewidth]{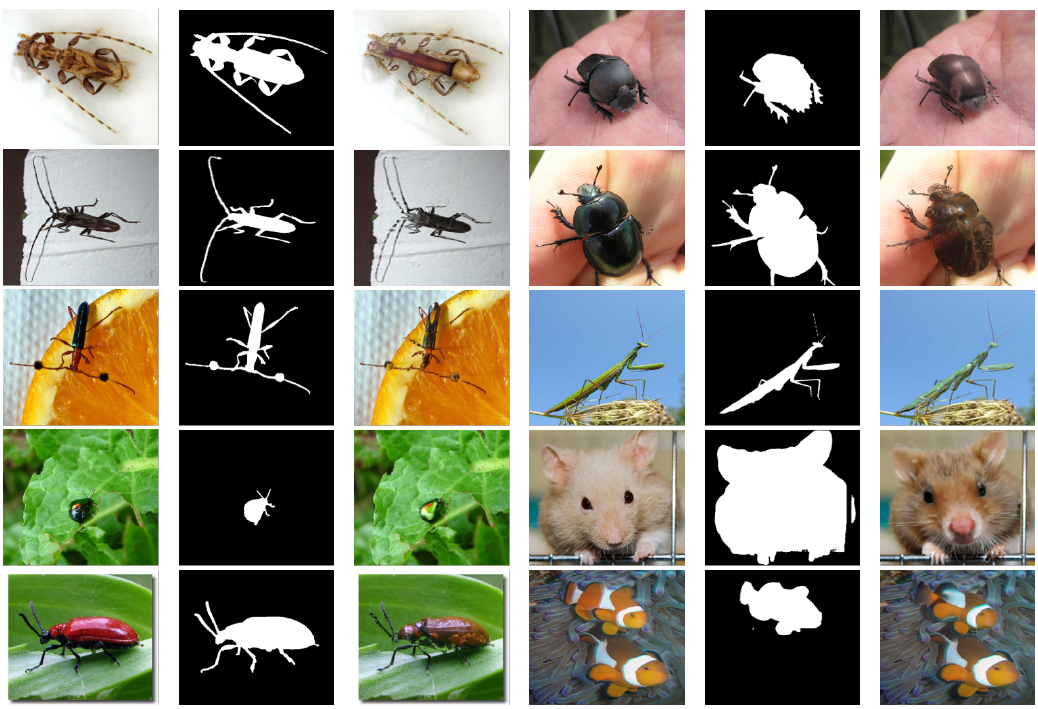}
	\caption{Additional visualizations of GIM-GLIDE. From left to right, the sequence is as follows: original image, tampered area, generative manipulated image, original image, tampered area, generative manipulated image.}
	\label{fig:GLIDE}
\end{figure*}

\begin{figure*}[!t]
	\centering
	\includegraphics[width=1.0\linewidth]{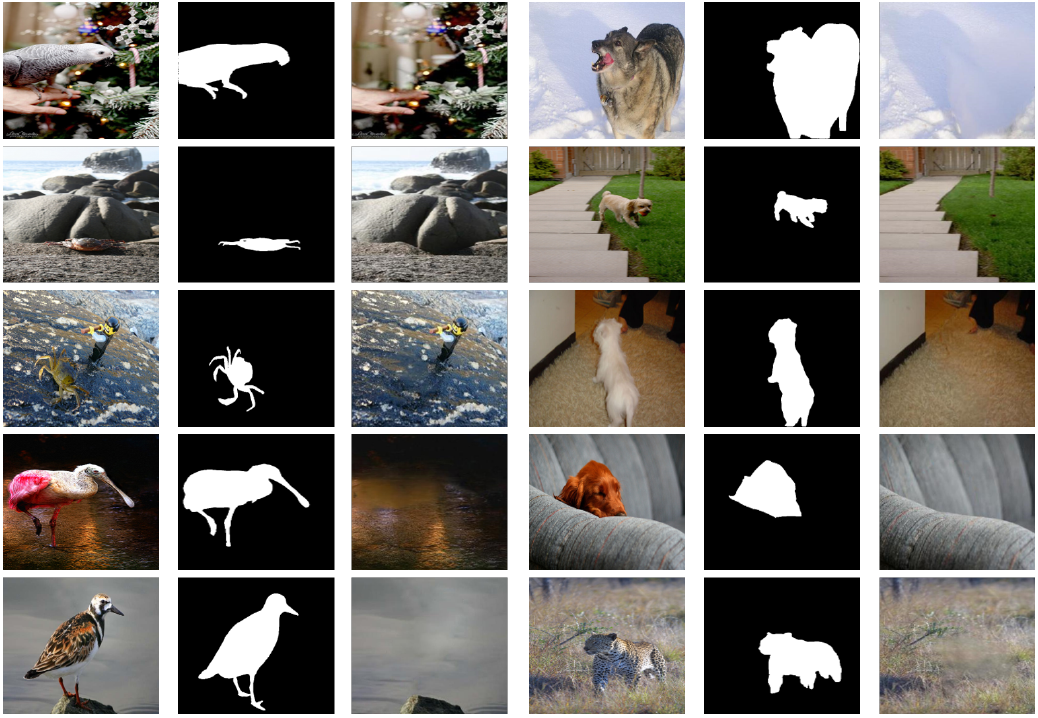}
	\caption{Additional visualizations of GIM-DDNM. From left to right, the sequence is as follows: original image, tampered area, generative manipulated image, original image, tampered area, generative manipulated image.}
	\label{fig:DDNM}
\end{figure*}

In this section, we present additional visualizations of the GIM dataset in Figure ~\ref{fig:SD}, ~\ref{fig:GLIDE}, and ~\ref{fig:DDNM}, showcasing the data visualization from GIM-SD, GIM-GLIDE, and GIM-DDNM respectively. 
These figures illustrate the diverse data representations within the GIM dataset, showcasing the unique characteristics brought by different generative models and manipulation techniques.

\section{Societal Impact}\label{sec:impact}

Our research contributes positively to society by addressing the challenge of detecting and locating generative-based manipulations, ultimately enhancing the security of AI-generated content (AIGC). The introduction of the reliable GIM database aims to bolster the effectiveness of Image Manipulation Detection and Localization (IMDL) methods, allowing for generalization across diverse scenarios. As a result, our algorithm GIMFormer instills greater trust among the general public in our society regarding media content authenticity. The utilization of GIM and GIMFormer proves advantageous for digital media forensics, particularly in addressing generative manipulation with real-world degradations. The practical applications of the GIM dataset have the potential to create a significant social impact. For instance, the training of detectors on GIM enables users to identify AI-generated content, leading to consequential actions such as identifying academic dishonesty. Widely disseminating and utilizing the GIM dataset for IMDL algorithms training could pave the way for its integration into specialized software designed for this purpose, thereby enhancing the accuracy and reliability of algorithms. However, it is essential to note that models trained on this dataset may exhibit undesirable tendencies towards novel objects in the real world, necessitating careful consideration for practical implementation despite their demonstrated generalizability.

\section{Ethics Statement}\label{sec:ethics}

Our GIM is based on ImageNet and VOC. No additional personally identifiable information or sensitive personally identifiable information is introduced during the production of fake images in the GIM dataset. During the dataset production, we do not introduce extra information containing exacerbated bias against people of a certain gender, race, sexuality, or who have other protected characteristics. The
ethical issues in the ImageNet and VOC datasets have been discussed in previous works. Crawford et al. ~\cite{crawford2021excavating} discuss issues with ImageNet. The first issue is the political nature of all taxonomies or classification systems, where terms like "male" and "female" are considered "natural," while "hermaphrodite" is offensively placed within the branch of Person > Sensualist > Bisexual alongside "pseudohermaphrodite" and "switch hitter" categories. The second issue concerns offensive images of real people, while the third is the use of people's photos without their consent by ImageNet creators.

\section{Limitations}
The GIM dataset shares classes with ImageNet and VOC, but it may not encompass future emerging objects due to the evolving variety in the real world. Fine-tuning pre-trained models on new object data can address this issue. Current research primarily centers on image manipulation. The emergence of realistic video generation models like SORA presents fresh challenges as AI manipulation extends into videos. Future plans involve expanding research to address the complexities of video manipulation.

\section{Acknowledgments}
This work is partially supported by NSFC ( No. 62402318, 62376153, 24Z990200676)
\bibliography{aaai25}

\end{document}